\documentclass[sigconf, authorversion, nonacm]{acmart}
\usepackage{bm, bbm}
\usepackage{algorithm, algorithmic}
\usepackage{multirow}
\usepackage[caption=false]{subfig}

\renewcommand{\algorithmicrequire}{\textbf{Input: }}
\renewcommand{\algorithmicensure}{\textbf{Output: }}

\newtheorem{lemma}{Lemma}

\AtBeginDocument{%
  }

\setcopyright{none}
\copyrightyear{}
\acmYear{}
\acmDOI{}

\acmConference[]{}{}
\acmPrice{}
\acmISBN{}

\begin{document}

\title{A Robust Classifier under Missing-Not-At-Random Sample Selection Bias}


\author{Huy Mai}
\email{huymai@uark.edu}
\affiliation{%
    \institution{University of Arkansas}
 \city{Fayetteville}
 \state{Arkansas}
 \country{USA}
 }

 \author{Wen Huang}
\email{wenhuang@uark.edu}
\affiliation{%
    \institution{University of Arkansas}
 \city{Fayetteville}
 \state{Arkansas}
 \country{USA}
 }

 \author{Wei Du}
\email{wd005@uark.edu}
\affiliation{%
    \institution{University of Arkansas}
 \city{Fayetteville}
 \state{Arkansas}
 \country{USA}
 }

\author{Xintao Wu}
\email{xintaowu@uark.edu}
\affiliation{%
    \institution{University of Arkansas}
 \city{Fayetteville}
 \state{Arkansas}
 \country{USA}
 }







\renewcommand{\shortauthors}{Mai et al.}

\begin{abstract}
  The shift between the training and testing distributions is commonly due to sample selection bias, a type of bias caused by non-random sampling of examples to be included in the training set. Although there are many approaches proposed to learn a classifier under sample selection bias, few address  the case where a subset of labels in the training set are missing-not-at-random (MNAR) as a result of the selection process. In statistics, Greene's method formulates this type of sample selection with logistic regression as the prediction model. However, we find that simply integrating this method into a robust classification framework is not effective for this bias setting. In this paper, we propose BiasCorr, an algorithm that improves on Greene's method by modifying the original training set in order for a classifier to learn under MNAR sample selection bias. We provide theoretical guarantee for the improvement of BiasCorr over Greene's method by analyzing its bias. Experimental results on real-world datasets demonstrate that BiasCorr produces robust classifiers and can be extended to outperform state-of-the-art classifiers that have been proposed to train under sample selection bias.
\end{abstract}


\settopmatter{printacmref=false}


\maketitle

\section{Introduction}


\noindent For the task of classification, the training and testing sets are generally assumed to be independently and identically distributed (IID), where the examples are drawn from the same distribution over covariates and labels. However, this assumption tends to be violated in the real-world. Dataset shift \cite{moreno2012unifying} describes the phenomenon in which the training and testing sets come from different distributions. One scenario that can cause dataset shift is sample selection bias, where an example is non-uniformly chosen from the population to be part of the training process. This type of bias can ultimately cause a set of training examples to be partially observed, where any of the covariates or label of an example is missing, or even completely unobserved. As a result, the performance of classifiers that are trained using a set subject to sample selection bias will be degraded. Most works have proposed solutions to problems dealing with missing-at-random (MAR) bias  \cite{cortes2008sample}, \cite{DBLP:conf/nips/HuangSGBS06}, \cite{liu2014robust}, \cite{DBLP:conf/icml/Zadrozny04}, where the non-inclusion of training samples is assumed to be independent from the label given the observed variables in the training set. However, these proposed solutions cannot properly account for the missing not at random (MNAR) setting, where the non-inclusion of training samples is assumed to not be independent from the label given the observed variables in the training set.

\begin{figure}[t]
    \centering
    \includegraphics[scale=0.40]{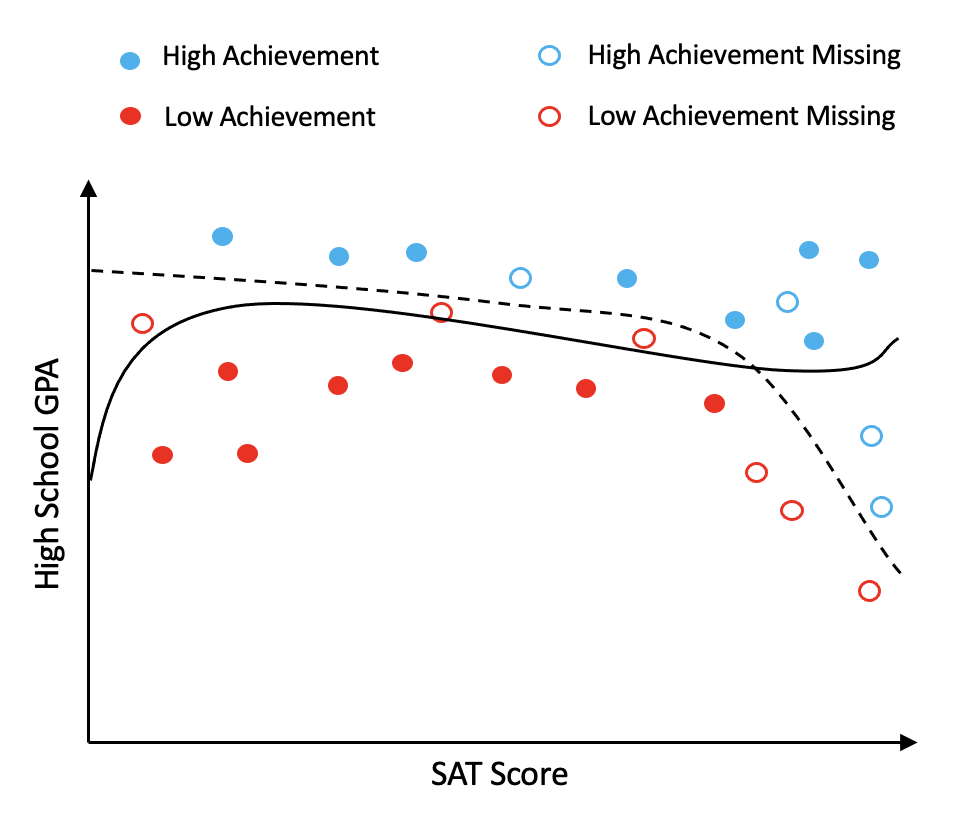}
    \caption{Illustration of the effect of MNAR sample selection bias on a classifier's predictions. Solid (dashed) line represents the decision boundary of the classifier trained on the biased (unbiased and fully observed) set.}
    \label{fig:illustrative}
\end{figure}

In this paper, we focus on MNAR sample selection bias on the labels. As an example of this type of bias, Figure \ref{fig:illustrative} shows the predicted achievement of an admitted freshman student based on their high school GPA and SAT score. The filled red and blue circles represent students with observed label values of achievement while the non-filled circles represent students with missing label values of achievement. There is some unknown mechanism about missing labels. For example, achievement values of students who have not declared their majors are not collected. When the undeclared students are omitted from the training, a biased model is produced and could be very different from the ground truth model that would have been trained had achievement of all students been collected. Our goal is to leverage the observed feature information of those records with missing labels in the training such that the trained model would be close to the ground truth model.

One classic method proposed to account for MNAR sample selection bias on the labels is Heckman's two-step method \cite{heckman1979sample}. Heckman's method utilizes inverse Mills ratio to model the impact of the selection process on the prediction. With the prediction and selection modeled as linear equations, this method constructs an unbiased model by first estimating the inverse Mills ratio (IMR) using the selection features and then incorporating it into the prediction equation. Due to its short computation time, Heckman's method has been a popular choice for solving linear regression under MNAR sample selection bias. However, applying Heckman's method to correct the same problem in the classification context is difficult. This is because the assumptions made for the use of the IMR may not be present in classifiers, causing them to perform inconsistently \cite{terza2009parametric}. 

To address this challenge, researchers turned their attention back to modeling the joint likelihood of prediction and selection \cite{miranda2006maxlikelihood, tucker2010selection}, which was a popular method before Heckman's method was proposed \cite{Amemiya1985advecon}. The joint likelihood considers the relationship between the noise terms used to model prediction and selection. In this work, we specifically examine the task of computing this likelihood using Greene's method \cite{greene2006general}. As a general framework for non-linear regression models under MNAR sample selection bias, Greene's method was one of the first methods to approximate the joint likelihood in order to reduce computational complexity. We find that although the method provides a first-order optimization process to produce an optimal solution, the minimization of its loss function over the biased training set does not take samples with missing labels into account. Thus, Greene's formulation alone cannot be used as an objective function when attempting to learn a classifier robust to MNAR sample selection bias.

\subsection{Problem Definition}

We give a summary of important symbols used throughout the paper in Table \ref{tab:notation}. Formally, let $\bm{\mathcal{X}}$ be the feature space and $\mathcal{Y}$ be the binary target attribute. We first consider the training set $\mathcal{D}_{tr}=\{t_i\}_{i=1}^n$ of $n$ samples that are originally sampled from the population to be modeled yet biased under MNAR sample selection bias. Each sample $t_i$ is defined as:
\begin{equation}
    t_i = \begin{cases}
    (\bm{x}_i,y_i,s_i=1) & 1\leq i\leq m \\
    (\bm{x}_i, s_i=0) & m+1\leq i\leq n
    \end{cases}
\end{equation} 
where the binary variable $s_i$ indicates whether or not $y_i$ is observed for a training sample. Let $\mathcal{D}_s$ denote the set containing the first $m$ training samples where each sample is fully observed and $\mathcal{D}_u$ be the set that contains the remaining $n-m$ training samples with unobserved labels.

\begin{table}[t]
    \small
    \centering
    \caption{Symbols.}
    \begin{tabular}{|c|c|}
    \hline
        \textbf{Notation} & \textbf{Description} \\
    \hline
        $h$ & robust classifier \\
        $\mathcal{D}_s$ ($\mathcal{D}_u$) & labeled (unlabeled) samples in training set $\mathcal{D}_{tr}$ \\
        $m (n)$ & size of $\mathcal{D}_s (\mathcal{D}_{tr})$ \\
        $\eta$ & missingness ratio in $\mathcal{D}_{tr}$ \\
        $\bm{x}_i$ & set of all features for $i$th training sample \\
        $y_i$ & label of $i$th training sample \\
        $s_i$ & selection value of $i$th training sample \\
        $\bm{x}_{i}^{(s)}$ ($\bm{x}_{i}^{(p)}$) & selection (prediction) features \\
        $\bm{\beta}$ & set of prediction coefficients \\
        $\hat{\mathcal{L}}$ & simulated negative log likelihood (Greene's) \\
        $\hat{\mathcal{L}}'$ & simulated negative log likelihood (BiasCorr) \\
        $R$ & number of random draws \\
        $\Bar{s}$ & estimated soft selection value \\
        $g_s$ ($g_y$) & predictor for selection (pseudolabel) \\
    \hline
    \end{tabular}
    \label{tab:notation}
\end{table}

We consider the following definitions to formally describe the MNAR sample selection bias scenario on $\mathcal{D}_{tr}$. We start with the MAR assumption: 

\begin{definition}
    \textbf{MAR Sample Selection}: \textit{ \underline{Missing-at-random} occurs for a sample $t_i$ if $s_i$ depends on $\bm{x}_i$ but is independent of $y_i$ given $x_i$, i.e. $P(s_i\vert \bm{x}_i, y_i) = P(s_i\vert \bm{x}_i)$ }.
\end{definition}

We then have the MNAR assumption, where MAR is violated:

\begin{definition}
    \textbf{MNAR Sample Selection}: \textit{ \underline{Missing-not-at-random} occurs for a sample $t_i$ if $s_i$ is not independent of $y_i$ given $x_i$, i.e. $P(s_i\vert \bm{x}_i, y_i) \neq P(s_i\vert \bm{x}_i)$ }.
\end{definition}

This means that $s_i$ may depend on $\bm{x}_i$ and $y_i$. For Greene's method, the selection mechanism is expressed in terms of a set of selection features to model the missingness of $y_i$ for a training sample. These selection features are observed for all training samples. Thus the following assumptions are additionally made in this work:
\begin{itemize}
    \item[(i)] Given a set of selection features $\bm{x}_{i}^{(s)}\subseteq \bm{x}_i$, $P(s_i\vert \bm{x}_i, y_i)$ is approximated by computing $P(s_i\vert \bm{x}_{i}^{(s)})$.
    \item[(ii)] The set of selection features includes every prediction feature, i.e. $\bm{x}_{i}^{(s)}\supset \bm{x}_{i}^{(p)}$.
\end{itemize}


\noindent \textbf{Problem Statement.} Given a set of prediction features $\bm{x}_{i}^{(p)}\subseteq \bm{x}_i$, we seek to train a binary classifier $h(\bm{x}_{i}^{(p)}; \bm{\beta})$ with parameters $\bm{\beta}$ that learns to minimize a loss function over biased training set $\mathcal{D}_{tr}$. 


\subsection{Contributions}

The contributions of our work are as follows. First, we propose BiasCorr, a framework for learning robust classification under MNAR sample selection bias, specifically in the case where the labels of some training samples are missing non-randomly. BiasCorr improves Greene's method by modifying the biased training dataset to assign a pseudolabel and an estimated soft selection value to the samples that have missing labels. These assignments are obtained by training two separate classifiers, one to predict pseudolabels by training on the set of fully observed training samples and another to predict the ground-truth selection value of a training sample. Second, we justify the improvement of BiasCorr over Greene's method by theoretically analyzing the bias of the loss functions estimated in BiasCorr and Greene's method. We provide theoretical guarantee to show that based on the level of missingness in the training set, the bias of BiasCorr is lower than that of Greene's method. Third, we extend BiasCorr to the classic problem of learning robust classification given a set of labeled training samples that are biased due to a hidden non-random selection process and an unbiased set of unlabeled samples. For the extension, we augment the training set with samples from the unbiased set, where the augmented samples are chosen by comparing the empirical frequencies of the biased training set and a set of samples randomly drawn from the unbiased set. Fourth, we conduct experiments on the real-world datasets and report the performance of our algorithms against state-of-the-art robust classifiers that were proposed under sample selection bias.

\section{Related Work}

\subsection{Regression under Sample Selection Bias}

Within the field of statistics, loss functions designed to solve the problem of modeling under MNAR bias have generally been categorized as full information maximum likelihood (FIML) estimators. FIML estimators estimate the coefficients of the prediction model while simultaneously accounting for the observation of labels for each sample. Each FIML estimator is centered around two linear equations: one that models the non-random selection of some data instance based on selection attributes and another that models the prediction of a value for a given data instance. The relationship between the two equations lies in the noise terms, which are assumed to be positively correlated in a bivariate normal distribution \cite{Amemiya1985advecon}.


Heckman's two-step method \cite{heckman1979sample} addresses the issue of sample selection bias within the context of linear regression when the dependent variable of a dataset has values that are MNAR. Unlike FIML estimators, this method requires estimating the inverse Mills ratio using the coefficients of the selection equation. The IMR is included as part of a new noise term for the prediction equation, namely the nonzero conditional noise expectation, in order to account for the bivariate normal relationship between the noise terms. Despite its popularity, Heckman's method has some key limitations when applied to non-linear regression models. One limitation of the method, noted by \cite{terza2009parametric}, relates to the nonzero conditional noise expectation term. For non-linear models, this term does not contain the IMR. For example, in the context of sample selection bias for Poisson regression, \cite{terza1998count} derived a term for the nonzero conditional noise expectation that included a nonlinear function similar in nature to the IMR, yet the IMR itself was not included. Moreover, the IMR may be incorrectly specified given the collinearity between the coefficients of the selection and prediction equations \cite{puhani2000heckman}. In the area of fair machine learning, \cite{du2021fairreg} formulated a fair regression model under the assumption that a subset of training outcomes are MNAR. The model, which has a closed-form solution under some fairness metric, adopts Heckman's method as part of its framework to account for the sample selection bias. Unlike these approaches, where the dependent variable is assumed to be continuous, our approach handles sample selection bias where the dependent variable is categorical. As closed-form solutions do not exist for likelihood equations maximized for logistic regression models, we depend on iterative optimization techniques in order to learn a classifier under MNAR sample selection bias.

\subsection{Classification under Sample Selection Bias}


Most research works in the area of learning under sample selection bias fall in the category of MAR bias. To address MAR sample selection bias, importance weighting techniques are often used, where training data instances are reweighted based on the ratio between the densities of the testing and training distributions \cite{bickel2007discriminative, DBLP:conf/icml/Zadrozny04}. With the possibility of these techniques resulting in inaccurate estimates due to the influence of data instances with large importance weights on the reweighted loss, other researchers incorporated ideas of minimax estimation to formulate models that are robust to MAR sample selection bias \cite{hu2018does, liu2014robust}. These models consider a worst-case shift between the training and testing distributions to adversarially minimize the reweighted loss. Approaches that handle MAR bias generally assume a labeled training set of biased samples and an unlabeled testing set of unbiased samples \cite{bickel2007discriminative}. As we address MNAR bias, we differ from these assumptions. In our study, we assume that the testing set cannot be accessed during training and that the training set contains a mixture of labeled and unlabeled examples given that the labels are non-randomly selected.

In recommender learning, \cite{wang2019doubly} proposed the joint learning of an imputation model and a prediction model to estimate the performance of rating prediction given MNAR ratings. \cite{lee2021dual} adopted two propensity scoring methods into its loss function to handle bias of MNAR implicit feedback, where user feedback for unclicked data can be negative or positive-yet-unobserved. While the approaches in \cite{lee2021dual, wang2019doubly} also use separate propensity estimation models to predict the observation of a label, they consider matrix factorization as the prediction model, which is not for binary classification on tabular data.

The problem we define in this work is related to semi-supervised learning \cite{vanengelen2020survey} where a training sample is treated differently based on whether the sample has a label or not. For labeled samples, the algorithm uses traditional supervision to update the model weights while for unlabeled samples, the algorithm often minimizes the difference in predictions between other similar training samples. In general, semi-supervised learning algorithms do not account for the missing data mechanism when comparing with similar samples. However, in our work, we model the missing data mechanism as we compare similar samples. One popular technique used in this setting is pseudolabel generation \cite{Lee2013PseudoLabelT}, where pseudolabels are made available to unlabeled samples based on predictions made by the model early in its training. This technique has been used to address MNAR labels. \cite{hu2022non} employed class-aware propensity score and imputation strategies using pseudolabels to develop a semi-supervised learning model that is doubly robust against MNAR data. This approach computes the probability of label missingness for a training sample in terms of a class prior. On the other hand, our approach does not require a class prior to compute the probability of label missingness for a training sample.  

\section{Greene's Method Revisited}

Greene's method \cite{greene2006general}  is an FIML estimator that accounts for the impact of non-random sample selection bias on the label by considering the relationship between the noise terms in the prediction and selection equations. Unlike Heckman's method, Greene's method estimates the variance and correlation coefficient of the noise terms as the likelihood of the model is maximized. This is important as the noise term of the prediction equation may have different distributions for various non-linear models (e.g., the noise term of the prediction equation for negative binomial regression has a log gamma distribution). As a result, Greene's method can be extended to non-linear regression cases.

\subsection{Sample Selection Model}

For any $(\bm{x}_i, y_i)\in \bm{\mathcal{X}}\times \mathcal{Y}$, the selection equation of the $i$th sample is $z_i = \bm{\gamma}\bm{x}_{i}^{(s)}+u_{i}^{(s)}$, where $\bm{\gamma}$ is the set of regression coefficients for selection, $\bm{x}_{i}^{(s)}$ is the set of features for sample selection, and $u_{i}^{(s)}\sim \mathcal{N}(0,1)$ is the noise term for the selection equation. The selection value of the $i$th sample $s_i$ is defined as:
\begin{equation}
s_i =\left\{
\begin{aligned}
&1 \hspace{0.5cm} z_i > 0 \\
&0  \hspace{0.5cm} z_i \le 0 \\
\end{aligned}
\right.
\end{equation} 

The prediction equation $f(y_i\vert \bm{x}_{i}^{(p)}, \epsilon_i)$ of the $i$th sample is based on logistic regression with
\begin{equation}
    f(y_i = 1\vert \bm{x}_{i}^{(p)}, \epsilon_i) = \frac{ \text{exp}(\bm{\beta}\bm{x}_{i}^{(p)}+\sigma\epsilon_i) }{ 1+ \text{exp}(\bm{\beta}\bm{x}_{i}^{(p)}+\sigma\epsilon_i) }
\label{eq: logistic regression}
\end{equation}
where $\bm{\beta}$ is the set of regression coefficients for prediction, $\bm{x}_{i}^{(p)}$ is the set of features for prediction, and $\sigma\epsilon_i$ is the noise term for the prediction equation, with $\sigma$ as the standard deviation of the term and $\epsilon_i\sim \mathcal{N}(0,1)$ as a random variable. We express $\sigma\epsilon_i$ as $u_{i}^{(p)}$, where $u_{i}^{(p)}\sim \mathcal{N}(0, \sigma^2)$.

The noise terms $u_{i}^{(s)}$ and $u_{i}^{(p)}$ are assumed to be bivariate normal, i.e. $u_{i}^{(s)} = \rho\epsilon_i + \sqrt{1 - \rho^2}v_i$, where $\rho$ is the correlation coefficient between $u_{i}^{(s)}$ and $u_{i}^{(p)}$ and $v_i\sim \mathcal{N}(0,1)$ is a random variable independent to $\epsilon_i$.

\subsection{Loss Function}

Based on the above sample selection model, the loss function
\begin{equation}
    \mathcal{L}= -\frac{1}{n} \sum_{i=1}^n \log\left( f(y_i, s_i\vert \bm{x}_{i}^{(p)}, \bm{x}_{i}^{(s)}) \right)
\end{equation}
over $\mathcal{D}_{tr}$ is then derived, which depends on the joint density function $f(y_i, s_i\vert \bm{x}_{i}^{(p)}, \bm{x}_{i}^{(s)})$. The first step is to consider $f(y_i, s_i = 1\vert \bm{x}_{i}^{(p)}, \bm{x}_{i}^{(s)})$, which is expressed as
\begin{equation}
    \small
    f(y_i, s_i = 1\vert \bm{x}_{i}^{(p)}, \bm{x}_{i}^{(s)}) = \int_{-\infty}^{\infty} f(y_i, s_i = 1\vert \bm{x}_{i}^{(p)}, \bm{x}_{i}^{(s)}, \epsilon_i)\cdot f(\epsilon_i)d\epsilon_i
\end{equation}

\begin{figure*}[h]
    \centering
    \includegraphics[scale=0.35]{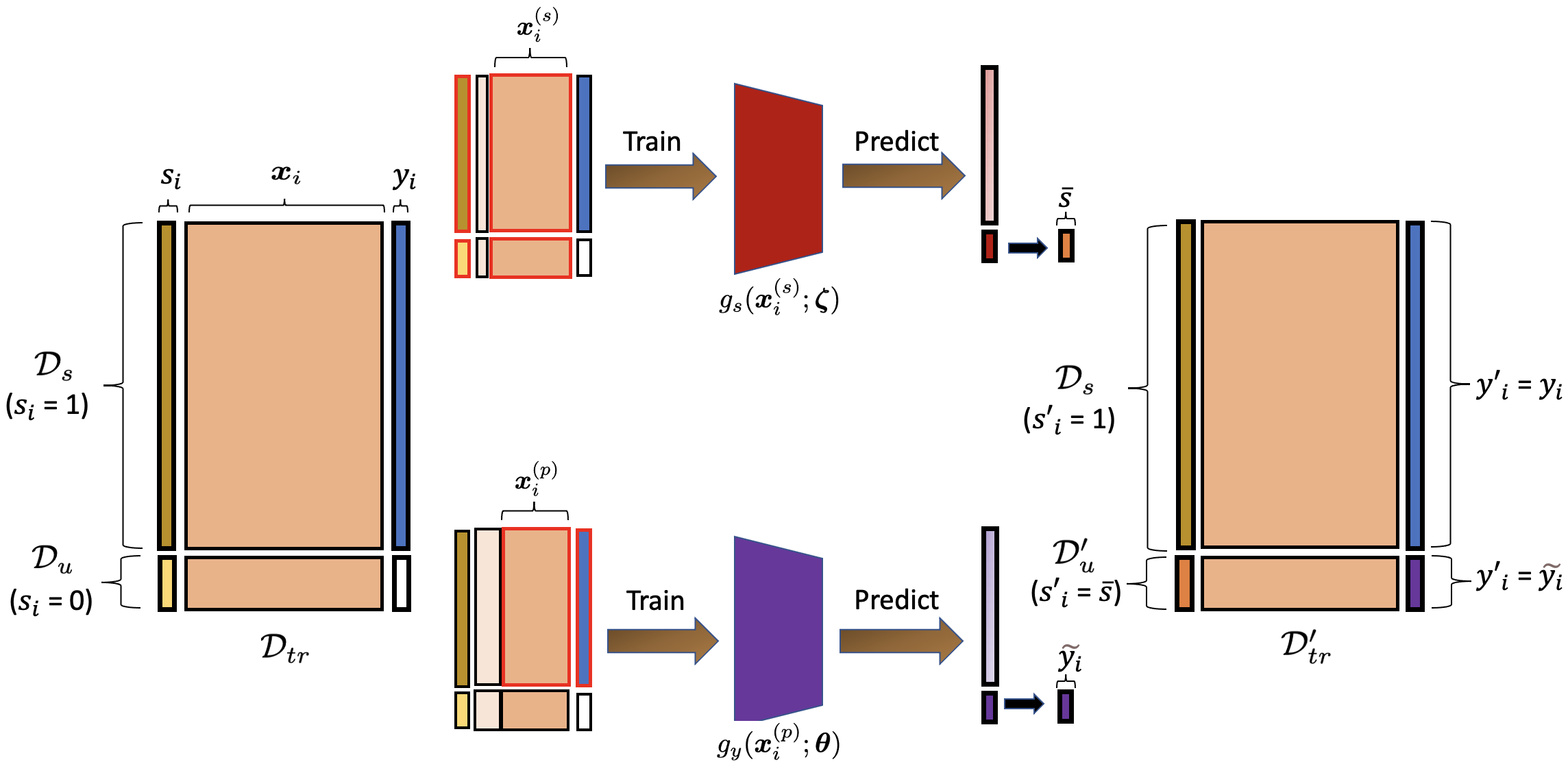}
    \caption{Process of producing $\mathcal{D}'_{tr}$ using BiasCorr. The boxes outlined in red indicate the parts of $\mathcal{D}_{tr}$ used to train $g_s$ and $g_y$.}
    \label{fig:biascorr}
\end{figure*}

\noindent Both $y_i$ and $s_i$ are independent when conditioned on $\epsilon_i$. Thus,
\begin{equation}
    \small
    f(y_i, s_i = 1\vert \bm{x}_{i}^{(p)}, \bm{x}_{i}^{(s)}, \epsilon_i) = f(y_i\vert \bm{x}_{i}^{(p)}, \epsilon_i)\cdot P(s_i = 1\vert \bm{x}_{i}^{(s)},\epsilon_i)
\end{equation}
Because $u_{i}^{(s)}$ and $u_{i}^{(p)}$ are bivariate normal,
we have
\begin{equation} \label{eq:ps1}
    P(s_i = 1\vert \bm{x}_{i}^{(s)},\epsilon_i) = \Phi\left( \frac{\bm{\gamma}\bm{x}_{i}^{(s)}+\rho\epsilon_i}{\sqrt{1 - \rho^2}} \right)
\end{equation}
where $\Phi(\cdot)$ is the standard normal cumulative distribution function. Since $\epsilon_i\sim \mathcal{N}(0,1)$,  $f(\epsilon_i)$ is $\phi(\epsilon_i)$, where $\phi(\cdot)$ is the standard normal density function. Thus,
\begin{multline} \label{eq:si1}
    \small
    f(y_i, s_i = 1\vert \bm{x}_{i}^{(p)}, \bm{x}_{i}^{(s)}) = \\
    \int_{-\infty}^{\infty} f(y_i\vert \bm{x}_{i}^{(p)}, \epsilon_i)\cdot \Phi\left( \frac{\bm{\gamma}\bm{x}_{i}^{(s)}+\rho\epsilon_i}{\sqrt{1 - \rho^2}} \right)\cdot \phi(\epsilon_i)d\epsilon_i
\end{multline}

\noindent The next step is to consider $f(y_i, s_i = 0\vert \bm{x}_{i}^{(p)}, \bm{x}_{i}^{(s)})$. The term $s_i = 0$ implies that the $y_i$ are missing. We have
\begin{equation}
    f(y_i, s_i = 0\vert \bm{x}_{i}^{(p)}, \bm{x}_{i}^{(s)}, \epsilon_i) = P(s_i = 0\vert \bm{x}_{i}^{(s)},\epsilon_i)
\end{equation}
where
\begin{equation} \label{eq:ps0}
    P(s_i = 0\vert \bm{x}_{i}^{(s)},\epsilon_i) = \Phi\left( -\frac{\bm{\gamma}\bm{x}_{i}^{(s)}+\rho\epsilon_i}{\sqrt{1 - \rho^2}} \right)
\end{equation}
So
\begin{equation} \label{eq:si0}
    \small
    f(y_i, s_i = 0\vert \bm{x}_{i}^{(p)}, \bm{x}_{i}^{(s)}) = \int_{-\infty}^{\infty} \Phi\left(- \frac{\bm{\gamma}\bm{x}_{i}^{(s)}+\rho\epsilon_i}{\sqrt{1 - \rho^2}} \right)\cdot \phi(\epsilon_i)d\epsilon_i
\end{equation}
Therefore, combining Eq. (\ref{eq:si1}) and Eq. (\ref{eq:si0}),
\begin{multline}
    \small
    f(y_i, s_i\vert \bm{x}_{i}^{(p)}, \bm{x}_{i}^{(s)}) = \\
    \int_{-\infty}^{\infty} [(1-s_i)+s_if(y_i\vert \bm{x}_{i}^{(p)},\epsilon_i)] \cdot  P(s_i\vert \bm{x}_{i}^{(s)}, \epsilon_i)\cdot \phi(\epsilon_i)d\epsilon_i
\end{multline}
where 
\begin{equation} 
    P(s_i\vert \bm{x}_{i}^{(s)}, \epsilon_i) = \Phi\bigg[(2s_i-1)\left(\frac{\bm{\gamma}\bm{x}_{i}^{(s)}+\rho\epsilon_i}{\sqrt{1 - \rho^2}}\right)\bigg]
\end{equation}
Thus the negative log likelihood function $\mathcal{L}$ over $n$ training data samples is
\begin{multline} \label{eq:logint1}
    \small
    \mathcal{L} = -\frac{1}{n}\sum_{i=1}^n \log\bigg( \int_{-\infty}^{\infty} [(1-s_i)+s_if(y_i\vert \bm{x}_{i}^{(p)},\epsilon_i)] \\
    \cdot P(s_i\vert \bm{x}_{i}^{(s)}, \epsilon_i)\phi(\epsilon_i)d\epsilon_i \bigg)
\end{multline}
$\mathcal{L}$ needs to be minimized with respect to $\bm{\beta}, \bm{\gamma}, \sigma,$ and $\rho$. Given that the computation of Eq. (\ref{eq:logint1}) is intractable, the simulation approach from \cite{train2009discrete} is used to minimize an approximate form of $\mathcal{L}$, denoted as $\hat{\mathcal{L}}$.
\begin{equation} \label{eq:greene}
    \hat{\mathcal{L}} = -\frac{1}{n}\sum_{i=1}^n \hat{l}_i
\end{equation}
where
\begin{equation}\label{eq:greenegrad}
\small
    \hat{l}_i = \log\bigg( \frac{1}{R}\sum_{r=1}^{R} [(1-s_i)+s_if(y_i\vert \bm{x}_{i}^{(p)},\epsilon_{ir})] \cdot P(s_i\vert \bm{x}_{i}^{(s)}, \epsilon_{ir}) \bigg)
\end{equation}
This approach involves taking $R$ random draws $\epsilon_{ir}$ from the standard normal population for the $i$th sample. As long as $R$ is greater than $\sqrt{n}$, then asymptotically $\hat{\mathcal{L}} = \mathcal{L}$. A proof of this claim is provided in \cite{lee1995asympt}.

\subsection{Optimization}
Iterative first-order optimization techniques such as stochastic gradient descent can be used to solve Eq. (\ref{eq:greene}) and obtain an optimal parameter $\bm{\beta}^*$ for the classifier $h$. We note that the gradient of Eq. (\ref{eq:greenegrad}) with respect to $\bm{\beta}$ for the $i$th training sample is expressed as
\begin{multline} \label{eq:gradient}
    \small
    \nabla_{\bm{\beta}}\hat{l}_i = \frac{1}{\hat{l}_i}\frac{1}{R}\sum_{r=1}^{R}s_i \cdot P(s_i\vert \bm{x}_{i}^{(s)}, \epsilon_{ir}) \\
    \cdot f(y_i\vert \bm{x}_{i}^{(p)},\epsilon_{ir})\cdot \bm{x}_{i}^{(p)} \cdot \frac{\partial f(y_i\vert \bm{x}_{i}^{(p)},\epsilon_{ir})}{\partial \bm{\beta}}
\end{multline}

\noindent We also apply the first-order optimization techniques to compute the other optimal parameters in Eq. (\ref{eq:greene}), namely $\bm{\gamma}^*$, $\sigma^*$, and $\rho^*$.

\section{Robust Classification under MNAR Sample Selection Bias}

Despite Greene's method incorporating a sample selection model towards fitting logistic regression, the task of training a robust classifier $h$ over $\mathcal{D}_{tr}$ under MNAR sample selection bias cannot be accomplished using this method. We specifically note a key issue in the optimization process. For any sample in the training set such that $s_i=0$, the value of Eq. (\ref{eq:gradient}) is 0, meaning that $\nabla_{\bm{\beta}}\hat{l}_i$ would account for only samples such that $y_i$ is observed. Thus, using a first-order optimization technique to solve Eq. (\ref{eq:greene}) does not result in an iterative solution $\bm{\beta}^*$ such that the classifier $h(\bm{x}_{i}^{(p)};\bm{\beta}^*)$ is robust against MNAR  sample selection bias on the label.

However, learning a robust classifier under MNAR sample selection bias can still be achieved by making improvements to Greene's method. First, we can refine the selection value of each sample in $\mathcal{D}_u$ to have a soft value in order to include information regarding the losses of samples in $\mathcal{D}_u$ when optimizing the classifier. While making the refinement, we still assume that each sample in $\mathcal{D}_s$ is assigned $s_i=1$. Second, we can impute the missing labels in $\mathcal{D}_u$ with pseudolabels to further improve Greene's method.

\begin{algorithm}[ht]
\caption{BiasCorr($g_s, g_y$)}\label{alg:cap}
\begin{flushleft}
    \algorithmicrequire Original training set $\mathcal{D}_{tr}=\{(\bm{x}_i, y_i, s_i=1)\}_{i=1}^{m}\cup \{(\bm{x}_i, s_i=0)\}_{i=m+1}^{n}$, $g_s$ model, $g_y$ model \\
    \algorithmicensure $\bm{\beta}^*$
\end{flushleft}
\begin{algorithmic}[1]
    \STATE $\mathcal{D}_s\gets \{(\bm{x}_i, y_i, s_i=1)\}_{i=1}^{m}$
    \STATE $\mathcal{D}_u\gets \{(\bm{x}_i, s_i=0)\}_{i=m+1}^{n}$
    \STATE $\mathcal{D}'_u\gets \emptyset$
    \STATE Train classifier $g_s(\bm{x}_{i}^{(s)}; \bm{\zeta})$ on $\mathcal{D}_{tr}$ to predict $s_i$
    \STATE Train classifier $g_y(\bm{x}_{i}^{(p)}; \bm{\theta})$ on $\mathcal{D}_s$ to predict $y_i$
    \FOR{$t_i$ in $\mathcal{D}_u$}
        \STATE $p^{(s)}_i\gets g_s(\bm{x}_{i}^{(s)}; \bm{\zeta})$
        \STATE $\tilde{y}_i\gets \mathbbm{1}[g_y(\bm{x}_{i}^{(p)}; \bm{\theta}) > 0.5]$
    \ENDFOR
    \STATE $\Bar{s}\gets \frac{1}{n-m}\sum_{i=m+1}^{n} p^{(s)}_{i}$
    \FOR{$i\in \{ m+1,\ldots, n \}$}
        \STATE $\mathcal{D}'_u\gets \mathcal{D}'_u\cup \{(\bm{x}_i, y'_i=\Tilde{y}_i, s'_i=\Bar{s})\}$
    \ENDFOR
    \STATE $\mathcal{D}'_{tr}\gets \mathcal{D}_s\cup \mathcal{D}'_u$
    \STATE Train $h(\bm{x}_{i}^{(p)}; \bm{\beta})$ to minimize $\hat{\mathcal{L}}'$ using $\mathcal{D}'_{tr}$ and obtain $\bm{\beta}^*$
\end{algorithmic}
\end{algorithm}

\subsection{BiasCorr}
\label{sec: biascorr algorithm}
To ensure that we learn classifiers that are robust to MNAR sample selection bias,  we introduce BiasCorr, a framework that addresses the challenge of training a classifier using Greene's method. In BiasCorr, we ensure that the losses of samples with missing labels are included in the optimization process. Using this framework, we train $h(\bm{x}_{i}^{(p)};\bm{\beta})$ to minimize $\hat{\mathcal{L}}'$, which is an enhanced version of Eq. (\ref{eq:greene}), over a modified training set $D'_{tr}$. We make these modifications while conforming to the original MNAR conditions on the label. Figure \ref{fig:biascorr} gives an illustration of the process to obtain $\mathcal{D}'_{tr}$.

Using the same assumptions as Greene's method on the training set $\mathcal{D}_{tr}$, BiasCorr assigns both an estimated soft selection value $\Bar{s}$ and a pseudolabel $\tilde{y}$ to each sample in $\mathcal{D}_u$, resulting in $h$ training to minimize the equation
\begin{equation}\label{eq:enhancedgreene} 
    \hat{\mathcal{L}}' = -\frac{1}{n}\sum_{i=1}^n \log\bigg( \frac{1}{R}\sum_{r=1}^{R} [(1-s'_i)+s'_if(y'_i\vert \bm{x}_{i}^{(p)},\epsilon_{ir})] 
    \cdot P(s'_i\vert \bm{x}_{i}^{(s)}, \epsilon_{ir}) \bigg)    
\end{equation}
over $\mathcal{D}'_{tr}=\mathcal{D}_s\cup \mathcal{D}'_u$, where
\begin{equation}
    s'_i = \begin{cases}
        1 & t_i\in \mathcal{D}_s \\
        \Bar{s} & t_i\in \mathcal{D}_u \\
    \end{cases}
\end{equation}
and
\begin{equation}
    y'_i = \begin{cases}
        y_i & t_i\in \mathcal{D}_s \\
        \tilde{y}_i & t_i\in \mathcal{D}_u \\
    \end{cases}
\end{equation}

To estimate the soft selection value $\Bar{s}$, we start by computing the probability $p^{(s)}_i$ of predicting $s_i=1$ for all samples in $\mathcal{D}_u$. This is based on our observation that the value of $P(s_i=1\vert \bm{x}^{(s)}_i, \epsilon_i)$ is not always equal to 0 for some tuple in $\mathcal{D}_u$, where the ground truth selection value is $s_i=0$. In our framework, we train a separate binary classifier $g_s(\bm{x}_{i}^{(s)}; \bm{\zeta})$ on $\mathcal{D}_{tr}$ to predict $s_i$ and obtain $p^{(s)}_i$ based on predictions using $\mathcal{D}_u$. We then get a fixed soft selection value by taking the average value of $p^{(s)}_i$ for all samples in $\mathcal{D}_u$.

While there is no difference in formulating the loss for each sample in $\mathcal{D}_s$ when comparing Eq. (\ref{eq:enhancedgreene}) to Eq. (\ref{eq:greene}), we alter the calculation of the loss for each sample in $\mathcal{D}_u$ in the following ways. First, based on the new binary assignment of $s'_i$, we compute $P(s'_i=\Bar{s}\vert \bm{x}_{i}^{(s)}, \epsilon_{ir})$, which is expressed as
\begin{equation}
    \text{\small $P(s_i=\Bar{s}\vert \bm{x}_{i}^{(s)}, \epsilon_i) = \Phi\bigg[(2\Bar{s}-1)\left(\frac{\bm{\gamma}\bm{x}_{i}^{(s)}+\rho\epsilon_i}{\sqrt{1 - \rho^2}}\right)\bigg]$}
\end{equation}
Compared to Eq. (\ref{eq:ps0}), the quantity $\small \frac{\bm{\gamma}\bm{x}_{i}^{(s)}+\rho\epsilon_{ir}}{\sqrt{1 - \rho^2}}$ is multiplied by $2\Bar{s} - 1$. As this adjusted calculation changes the optimization of the selection coefficients $\bm{\gamma}$, we see that the value of $P(s'_i=1\vert \bm{x}_{i}^{(s)}, \epsilon_{ir})$, which is computed using Eq. (\ref{eq:ps1}), is greater than $P(s_i=1\vert \bm{x}_{i}^{(s)}, \epsilon_{ir})$ after $\hat{\mathcal{L}}'$ is minimized. As the formulation for computing the loss of each sample in $\mathcal{D}_s$ is kept fixed, this means that the training of $h$ using $\hat{\mathcal{L}}$ is improved as $\hat{\mathcal{L}}'$ is expected to converge to a value less than $\hat{\mathcal{L}}$.

Second, in Eq. (\ref{eq:enhancedgreene}), we have the term $[(1-\Bar{s})+\Bar{s}f(\tilde{y}_i\vert \bm{x}_{i}^{(p)},\epsilon_{ir})]$ which is multiplied by $P(s'_i=\Bar{s}\vert \bm{x}_{i}^{(s)}, \epsilon_{ir})$. This term can be interpreted as a weight that puts more emphasis on the label information of samples in $\mathcal{D}_u$ as the value of $\Bar{s}$ increases. Nevertheless, as we incorporate $\Bar{s}$ as a selection value assignment, it is impossible to compute $f(y_i\vert \bm{x}_{i}^{(p)}, \epsilon_{ir})$ for samples in $\mathcal{D}_u$ because of their missing labels due to MNAR bias. This is why we consider the prediction of pseudolabels $\tilde{y}_i$ when computing the loss for each sample in $\mathcal{D}'_u$.

The pseudocode for BiasCorr is provided in Algorithm \ref{alg:cap}. In line 4, we first train $g_s$ on $\mathcal{D}_{tr}$ to predict the original ground-truth selection value $s_i$. In line 5, we train another binary classifier $g_y(\bm{x}_{i}^{(p)}; \bm{\theta})$ with parameters $\bm{\theta}$ on $\mathcal{D}_s$ to predict the ground-truth label $y_i$. To add samples to $\mathcal{D}'_u$, in line 7, we evaluate $g_s$ and obtain the probability $p^{(s)}_i$ for each sample in $\mathcal{D}_u$. In line 8, we use the prediction from the evaluation of $g_y$ on each sample in $\mathcal{D}_u$ to obtain a pseudolabel $\tilde{y}_i$. In line 10, we compute the average $\Bar{s}$ of $p^{(s)}_i$ of each sample in $\mathcal{D}_u$. In line 12, we add each tuple $(\bm{x}_i, \Tilde{y}_i, \Bar{s})$ to $\mathcal{D}'_u$, where each $\bm{x}_i$ is taken from $\mathcal{D}_u$. In line 15, using $\mathcal{D}'_{tr}$ we obtain a solution $\bm{\beta}^*$ after minimizing Eq. (\ref{eq:enhancedgreene}) such that $h(\bm{x}_{i}^{(p)};\bm{\beta}^*)$ is robust against non-random sample selection bias on the label.

The computational complexity of Algorithm \ref{alg:cap} trivially depends on the complexity of training $g_s$, $g_y$, and $h$ to convergence. Similar to the training of $h$ using Eq. (\ref{eq:greene}), the complexity of training $h$ by minimizing Eq. (\ref{eq:enhancedgreene}) is $O(Tn)$, where $T$ is the number of training iterations for $h$.

We further note that the types of models used to train $g_s$ and $g_y$ are listed as inputs to Algorithm \ref{alg:cap}. In our work, we experiment training $g_s$ using the probit and logistic regression models. Compared to logistic regression models, which is based on the sigmoid function, probit models use the normal cumulative distribution function to model binary classification. For $g_y$, we consider logistic regression and multi-layer perceptron.

\subsection{Bias Analysis Regarding Loss Function}
\label{sec: bias analysis}
In this section, we analyze the bias of the loss function estimator for both Greene's method and BiasCorr algorithm. We compare the two biases and show that our BiasCorr algorithm further reduces the bias for classification performance estimation given the ratio of the unlabeled training set is larger than a threshold. We first define the optimized negative log-likelihood loss function where the training data $
\mathcal{D}_{tr}$ is fully observed:
\begin{equation}
\mathcal{L}^* = -\frac{1}{|\mathcal{D}_{tr}|} \sum_{i \in \mathcal{D}_{tr}} \log P(y_i|\bm{x}_i )  =  -\frac{1}{n} \sum_{i=1}^n \log\left( f(y_i \vert \bm{x}_{i}^{(p)}) \right)     
\end{equation}
where $f(y_i \vert \bm{x}_{i}^{(p)})$ takes the form of logistic regression in our paper. The bias of an arbitrary loss function estimator $\mathcal{L}$ is defined as:
\[ \textit{Bias}(\mathcal{L}) = \big|\mathcal{L}^* - \mathbb{E}_{\mathcal{D}_{tr}}[\mathcal{L}] \big|\]

Following previous definitions we have $|\mathcal{D}_{tr}| = n, |\mathcal{D}_{s}| = m$, we further define the missing ratio of the unlabeled training samples $|\mathcal{D}_{u}| / |\mathcal{D}_{tr}|$ as $\eta = 1 - \frac{m}{n}$. Denoting $p(s_i)$ as the ground truth selection probability for each tuple $i$ based on its selection features $\bm{x}_{i}^{(s)}$, and the expectation of the estimated selection model $P(s_i\vert \bm{x}_{i}^{(s)}, \epsilon_{ir})$ and prediction model $f(y_i\vert \bm{x}_{i}^{(p)},\epsilon_{ir})$ over $R$ random draws on the error terms as $\hat{p}(s_i)$ and  $\hat{f}(y_i\vert \bm{x}_{i}^{(p)})$ respectively.
We next formally derive the bias of the loss function estimators from Greene's method and BiasCorr algorithm in the following two lemmas:

\begin{lemma}[Bias of Greene's method estimator]
    Given the estimated selection model $\hat{p}(s_i)$ and outcome  model $ \hat{f}(y_i\vert \bm{x}_{i}^{(p)})$, the bias of the loss function estimator for Greene's method shown in Eq. (\ref{eq:greene}) is:
\begin{equation}\small
\textit{Bias}(\hat{\mathcal{L}}) = \bigg |  \frac{1}{n}\sum_{i=1}^n  \log \bigg( \frac{ f(y_i\vert \bm{x}_{i}^{(p)})}{  \hat{p}(s_i) + p(s_i)\hat{p}(s_i)( \hat{f}(y_i\vert \bm{x}_{i}^{(p)}) -1 ) }  \bigg)  \bigg|
\label{eq: bias_greene}
\end{equation}
\label{lemma: bias_greene}
\end{lemma}

\begin{lemma}[Bias of BiasCorr estimator]
Given the definitions of $s'_i,\Bar{s},y'_i$ in Section \ref{sec: biascorr algorithm}, the bias of the loss function estimator for BiasCorr algorithm  shown in Eq. (\ref{eq:enhancedgreene}) is:
\begin{equation}\small
      \textit{Bias}(\hat{\mathcal{L}}') =  \bigg |  \frac{1}{n}\sum_{i=1}^n  \log \bigg( \frac{ f(y_i\vert \bm{x}_{i}^{(p)})}{  \hat{p}(s'_i) + (p(s_i)+\Bar{s} \eta )\hat{p}(s'_i)( \hat{f}(y'_i\vert \bm{x}_{i}^{(p)}) -1 ) }  \bigg)  \bigg|
\label{eq: bias_biascorr}
\end{equation} 
\label{lemma: bias_biascorr}
\end{lemma}

Note that both $\textit{Bias}(\hat{\mathcal{L}})$ and $\textit{Bias}(\hat{\mathcal{L}'})$ are non-zero even if the estimated selection variable model and outcome model are accurate, that is, $\hat{p}(s_i) = p(s_i)$ and $ \hat{f}(y_i\vert \bm{x}_{i}^{(p)}) = f(y_i\vert \bm{x}_{i}^{(p)})$.  According to the design of the log-likelihood loss function in Eq. (\ref{eq:logint1}), Greene's method uses $\hat{f}(y_i\vert \bm{x}_{i}^{(p)}) \cdot \hat{p}(s_i)$ to estimate the likelihood function $f(y_i,s_i\vert \bm{x}_{i}^{(p)}, \bm{x}_i^{(s)})$ for the samples in $\mathcal{D}_{s}$ and $\hat{p}(s_i)$  to estimate  $f(y_i,s_i\vert \bm{x}_{i}^{(p)}, \bm{x}_i^{(s)})$ for the samples in $\mathcal{D}_{u}$. Due to the fundamental difference between selection model and prediction model, it is very challenging to derive an unbiased estimator for the loss function based on Greene's method. However, by applying the modification from BiasCorr algorithm, we are able to further reduce the bias for the loss function estimator on classification tasks based on an assumption on the ratio $\eta$.
We list our main theorem that compares the biases of the two methods as follows:

\begin{theorem}
Given a training dataset with labeled and unlabeled tuples $\mathcal{D}_{tr} = \mathcal{D}_s \cup \mathcal{D}_u$, suppose $\hat{f}(y_i\vert \bm{x}_{i}^{(p)})$  takes the form of logistic regression and there is no bias caused by the estimated selection model for both Greene's method and BiasCorr. If the ratio of the unlabeled training data $\eta$ is larger than $1/(2-\bar{s})$, we have  
\[\textit{Bias}(\hat{\mathcal{L}'}) < \textit{Bias}(\hat{\mathcal{L}})\]   
\label{thm: bias_diff}
\end{theorem}
To obtain the result in Theorem \ref{thm: bias_diff} we consider the difference between the two biases and analyze the terms after subtracting $\textit{Bias}(\hat{\mathcal{L}})$ by $\textit{Bias}(\hat{\mathcal{L}'})$. We first decompose the difference and derive the inequality as follows: 
\begin{equation}
\begin{split}
&\textit{Bias}(\hat{\mathcal{L}})- \textit{Bias}(\hat{\mathcal{L}'}) \geq  \\ &\Bar{s}\eta \cdot \bigg[  \underbrace{1 -  \Bar{s}\eta - \frac{1}{n} \sum_{i=1}^n 2p(s_i)}_\text{term 1} 
+ \underbrace{\big[ \frac{1}{n} \sum_{i=1}^n \hat{f}(y_i\vert \bm{x}_{i}^{(p)}) (2p(s_i) + \Bar{s}\eta )}_\text{term 2} \big] \bigg]
\end{split}
\label{eq: diff_inequality}
\end{equation}
According to Eq. (\ref{eq: diff_inequality}) we find that if both term 1 and term  2 are greater than 0, the BiasCorr estimator is guaranteed to achieve lower bias than Greene's model estimator. Since both $\hat{f}(y_i\vert \bm{x}_{i}^{(p)})$ and $p(s_i)$ lie in $(0,1)$ for each tuple $i$, term 2 is still a positive value after summation and averaging over all the training tuples. Our theoretical analysis also shows that to guarantee the positivity of term 1, the proportion of the unlabeled training data $\eta$ needs to be larger than  $1/(2-\bar{s})$. Notice that the condition $\eta \leq 1/(2-\bar{s})$ does not necessarily imply $\textit{Bias}(\hat{\mathcal{L}'})$ is larger than $\textit{Bias}(\hat{\mathcal{L}})$. We still need to compare the magnitude of term 1 and term 2, and the value of term 2 heavily depends on the estimated selection model and outcome model. Finally, combining the analysis results leads to the conclusion in Theorem \ref{thm: bias_diff}. For the proof details of Lemma \ref{lemma: bias_greene}, Lemma \ref{lemma: bias_biascorr}  and Theorem \ref{thm: bias_diff}, please refer to Appendix \ref{app: proof}. 

\subsection{Extending BiasCorr}

Most algorithms that have been proposed to learn classification under sample selection bias are trained under the assumption that the training set contains labeled samples $\mathcal{D}_s=\{(\bm{x}_i, y_i)\}_{i=1}^{m}$ that come from a biased source distribution. Additionally, they assume that there exists a set of testing samples $\mathcal{D}_N=\{\bm{x}_i\}_{i=1}^{N}$ drawn from an unbiased target distribution. We propose an extension of BiasCorr, BiasCorr$^*$, for this setting. We do so by augmenting the original set of labeled training samples using the set of unlabeled samples from the target distribution. Specifically, given two sets $\mathcal{D}_s$ and $\mathcal{D}_N$, we construct an augmented training set $\mathcal{D}_{aug}=\mathcal{D}_s\cup \mathcal{D}_u$ of $n$ samples, where $\mathcal{D}_u$ contains samples that are uniformly drawn from $\mathcal{D}_N$ and $n>m$.

\begin{algorithm}[h]
\caption{BiasCorr$^*$ ($g_s$, $g_y$)}\label{alg:biascorrextend}
\begin{flushleft}
    \algorithmicrequire Labeled training set $\mathcal{D}_s=\{(\bm{x}_i, y_i)\}_{i=1}^m$, unlabeled testing set $\mathcal{D}_N=\{(\bm{x}_i)\}_{i=1}^N$, $g_s$ model, $g_y$ model \\
    \algorithmicensure $\bm{\beta}^*$
\end{flushleft}
\begin{algorithmic}[1]
    \STATE $\mathcal{D}_n\gets$ $n$ randomly drawn samples from $\mathcal{D}_N$
    \STATE $\mathcal{D}_u\gets \emptyset$
    \FOR{each distinct sample $t$ in $\mathcal{D}_n$}
        \STATE $a_t\gets |\mathcal{D}^t_s|$
        \STATE $b_t\gets |\mathcal{D}^t_n|$
        \IF{$b_t>a_t$}
        \STATE Draw $b_t-a_t$ samples from $\mathcal{D}^t_n$ and add samples to $\mathcal{D}_u$
        \ENDIF
        \IF{$|\mathcal{D}_u|=n-m$}
            \STATE \textbf{break}
        \ENDIF
    \ENDFOR
    \STATE Assign $s_i=1$ (0) to all $t_i\in \mathcal{D}_s$ $(\mathcal{D}_u)$
    \STATE $\mathcal{D}_{aug}\gets \mathcal{D}_s\cup \mathcal{D}_u$
    \STATE $\bm{\beta}^*\gets$ \textit{BiasCorr}($\mathcal{D}_{aug}$, $g_s$, $g_y$)
\end{algorithmic}
\end{algorithm}

\begin{table*}[h]
    \small
    \centering
    \caption{Performance of baselines compared to BiasCorr. Highest test accuracies among SSBias, Greene's method, and the four BiasCorr settings are in bold.}
    \begin{tabular}{|c|c|c|c|c|c|c|}
    \hline
        \multirow{2}{*}{\textbf{Methods}} & \multicolumn{2}{c|}{\textbf{Adult}} & \multicolumn{2}{c|}{\textbf{German}} & \multicolumn{2}{c|}{\textbf{Drug}} \\ \cline{2-7} 
        & Train Acc. (\%) & Test Acc. (\%) & Train Acc. (\%) & Test Acc. (\%) & Train Acc. (\%) & Test Acc. (\%) \\
    \hline 
        NoBias & 86.57 $\pm$ 0.00 & 86.57 $\pm$ 0.00 & 73.29 $\pm$ 0.00 & 72.67 $\pm$ 0.00 & 69.83 $\pm$ 0.00 & 69.08 $\pm$ 0.00 \\
        SSBias & 77.56 $\pm$ 0.00 & 62.44 $\pm$ 0.00 & 75.28 $\pm$ 0.00 & 69.33 $\pm$ 0.00 & 77.78 $\pm$ 0.00 & 66.78 $\pm$ 0.00 \\
        Greene's method & 62.94 $\pm$ 0.07 & 62.89 $\pm$ 0.09 & 72.77 $\pm$ 0.47 & 69.67 $\pm$ 0.30 & 68.89 $\pm$ 0.27 & 66.71 $\pm$ 0.33 \\
        BiasCorr (probit, LR) & 86.84 $\pm$ 0.02 & 70.05 $\pm$ 0.04 & 79.97 $\pm$ 0.14 & \textbf{71.60 $\pm$ 0.13} & 87.93 $\pm$ 0.07 & \textbf{69.22 $\pm$ 0.02} \\
        BiasCorr (LR, LR) & 87.36 $\pm$ 0.04 & 69.84 $\pm$ 0.04 & 80.11 $\pm$ 0.25 & 71.07 $\pm$ 0.13 & 88.89 $\pm$ 0.09 & 67.81 $\pm$ 0.17 \\
        BiasCorr (probit, MLP) & 94.08 $\pm$ 0.02 & 85.68 $\pm$ 0.01 & 79.69 $\pm$ 0.40 & 71.27 $\pm$ 0.25 & 86.19 $\pm$ 0.06 & 67.39 $\pm$ 0.09 \\
        BiasCorr (LR, MLP) & 93.45 $\pm$ 0.01 & \textbf{85.79 $\pm$ 0.02} & 79.69 $\pm$ 0.50 & 71.00 $\pm$ 0.21 & 85.97 $\pm$ 0.13 & 67.77 $\pm$ 0.14 \\
    \hline
    \end{tabular}
    \label{tab:case1resultslr}
\end{table*}


Algorithm \ref{alg:biascorrextend} gives the pseudocode of BiasCorr$^*$. To obtain $\mathcal{D}_{aug}$, we first randomly draw $n$ samples uniformly from $\mathcal{D}_N$ in line 1, where $n>m$. Let $\mathcal{D}_n$ denote this set of $n$ samples\footnote{We note that in some cases, $N<m$. To obtain $D_n$, we draw $n$ samples from $\mathcal{D}_N$ with replacement.}. To construct $\mathcal{D}_u$, we compare the empirical frequencies of $\mathcal{D}_s$ and $\mathcal{D}_n$ in lines 4-8, which follows a similar procedure as \cite{cortes2008sample}. For a distinct sample $t$, let $\mathcal{D}_s^t$ be a subset of $\mathcal{D}_s$ that contains all instances of $t$ and $a_t=|\mathcal{D}_s^t|$. We similarly define $\mathcal{D}_n^t$ and $b_t$ for $\mathcal{D}_n$. Until $\mathcal{D}_u$ contains $n-m$ samples, we add $b_t - a_t$ random samples from $\mathcal{D}_n^t$ to $\mathcal{D}_u$ for each $t$ such that $b_t>a_t$. 

We note that choosing $n$ as the size of $\mathcal{D}_{aug}$ is significant in determining the performance of estimating the selection probability and the efficiency of BiasCorr$^*$. First, the following lemma from \cite{cortes2008sample} shows the error of using $\frac{a_t}{b_t}$ as an estimate of the selection probability $P(s_i=1\vert t)$.
\begin{lemma}
\cite{cortes2008sample} Let $\delta>0$. Let $a'$ be number of distinct samples in $\mathcal{D}_s$ and $p_0=\underset{t\in \mathcal{D}_{aug}}{min}P(t)\neq 0$. Then, with probability at least $1-\delta$, the following inequality holds for all distinct $t\in \mathcal{D}_s$:
\begin{equation}
    \bigg\vert P(s_i=1\vert t) - \frac{a_t}{b_t} \bigg\vert \leq \sqrt{ \frac{\log 2a' + \log \frac{1}{\delta}}{p_0n} }
\end{equation}
\end{lemma}

Here we see that for a given number of distinct samples in $\mathcal{D}_s$, the error of estimating $P(s_i=1\vert t)$ depends on the value of $p_0n$, which equals the number of occurrences of the least frequent sample in $\mathcal{D}_{aug}$. This value is dependent on the set $\mathcal{D}_u$, which may include samples $t$ that are not in $\mathcal{D}_s$. Second, the computational complexity of generating $\mathcal{D}_{aug}$ in lines is bounded by $n$, where in the worst case $\mathcal{D}_n$ has $n$ distinct samples and the last $n-m$ samples in $\mathcal{D}_n$ are added to $\mathcal{D}_u$.

\section{Experiments}

\subsection{Datasets}
We evaluate the performance of our proposed algorithms on the Adult, German, and Drug datasets \cite{adult} using the set of prediction and selection features listed in Table \ref{tab:features} in Appendix \ref{sec:features}. The Adult dataset contains 45,222 samples that were collected from the 1994 Census database. For each sample, we predict whether the type of workclass is government or private. The German dataset, which consists of attributes such as credit history, age, duration in months, and checking account status, contains 1,000 samples of individuals that have either good or bad credit risk. The task is to predict the level of credit risk for each person. The Drug dataset has 1,885 samples of respondents with known attributes such as gender, education, and ethnicity. Based on these attributes, we predict how likely it is for a respondent to use benzos. 

\subsection{Experiments on BiasCorr}

We choose 70\% of samples in each dataset to generate the original training set $\mathcal{D}_{tr}$. We work with two different bias scenarios in our experiments: one where the condition of the missingness ratio $\eta$ listed in Theorem \ref{thm: bias_diff} is satisfied and another where the condition is not met. To create the sample selection bias on $\mathcal{D}_{tr}$ for the Adult dataset, we select a training sample to have an observed label if the years of education is more than 12. As a result, out of 31,655 training samples, the set $\mathcal{D}_u$ contains 23,664 samples ($\eta=0.7476$). For the German dataset, we select a training sample to be fully observed if the person has been employed for more than 1 year, resulting in $\mathcal{D}_u$ having 162 out of 700 training samples ($\eta=0.2314$). For the Drug dataset, we create the sample selection bias scenario for the training set by selecting individuals whose Oscore is at most 43. As a result, 860 out of 1,319 samples are in $\mathcal{D}_u$ ($\eta=0.6520$). \\
\noindent \textbf{Baselines and Reproducibility.} We compare BiasCorr to the following baselines: (a) logistic regression without sample selection bias (NoBias), which is trained using  $\mathcal{D}_{tr}=\mathcal{D}_s\cup \mathcal{D}_u$ where all samples in $\mathcal{D}_{tr}$ are fully observed, (b) logistic regression with sample selection bias (SSBias), which is trained using $\mathcal{D}_s$, and (c) logistic regression with sample selection bias correction based on Greene's method, which is trained using the set $\mathcal{D}_s\cup \mathcal{D}_u$ where all samples in $\mathcal{D}_u$ have non-randomly missing labels. Our models and all baselines are implemented using Pytorch. Details regarding implementation and the hyperparameters used for BiasCorr are given in Appendix \ref{sec: hyperparam}. Our {\bf source code} can be downloaded using the link \url{https://tinyurl.com/4kvux87n}. \\
\noindent \textbf{Results.} Table \ref{tab:case1resultslr} shows the training/testing accuracy of each model. We report average accuracies and their standard deviations over 5 runs. We first see that while the change in training accuracies is different for each dataset when comparing NoBias and SSBias, NoBias outperforms SSBias by 24.13\%, 3.34\%, and 2.30\% when considering the testing accuracy for the Adult, German, and Drug datasets, respectively. This shows that the utility of the logistic regression model is reduced when trained on $\mathcal{D}_s$. 
We also see that Greene's method does not outperform SSBias by much when evaluated on the testing set. For instance, when looking at the results for the Adult dataset in Table \ref{tab:case1resultslr}, the testing accuracy of Greene's method is 0.55\% higher than SSBias while the testing accuracy of NoBias is 24.13\% higher. This demonstrates that a classifier is not robust to MNAR sample selection bias when learning to optimize Eq. (\ref{eq:greene}).


\begin{table}[t]
    \small
    \centering
    \caption{Empirical missingness ratio comparison.}
    \begin{tabular}{|c|c||c|c|}
    \hline
      \textbf{Dataset} & $\eta$ & $1/(2-\Bar{s})$ (probit for $g_s$) & $1/(2-\Bar{s})$ (logit for $g_s$) \\
    \hline
        Adult & 0.7476 & 0.5868 & 0.5738 \\
        German & 0.2314 & 0.6345 & 0.6233 \\
        Drug & 0.6520 & 0.7159 & 0.5976 \\
    \hline
    \end{tabular}%
    \label{tab:estimatedsbar}
\end{table}

\begin{table}[t]
    \small
    \centering
    \caption{Execution times (in seconds).}
    \begin{tabular}{|c|c|c|c|}
    \hline
        \textbf{Method} & Adult & German & Drug \\
    \hline
        Greene's method & 93.53 & 2.06 & 3.14 \\
        BiasCorr (probit, LR) & 94.59 & 1.84 & 2.85 \\
        BiasCorr (LR, LR) & 99.62 & 1.87 & 2.28 \\
        BiasCorr (probit, MLP) & 112.17 & 1.86 & 2.69 \\
        BiasCorr (LR, MLP) & 112.90 & 1.87 & 2.26 \\
    \hline
    \end{tabular}%
    \label{tab:execution}
\end{table}

More importantly, we observe that BiasCorr, under all 4 combinations of settings for $g_s$ and $g_y$, outperforms SSBias and Greene's method. Using the German dataset as an example, BiasCorr(LR, MLP) has the lowest test accuracy out of the four BiasCorr settings after training on the dataset. Despite this, BiasCorr(LR, MLP) outperforms SSBias by 1.67\% on the testing set. This difference is higher than the 0.34\% margin when comparing Greene's method to SSBias. 

We also examine the values of $\eta$ and $1/(2-\Bar{s})$ in Table \ref{tab:estimatedsbar} based on this experiment. Using $g_s$ on probit as an example, we see that the value of $1/(2-\Bar{s})$ is 0.5868 for the Adult dataset. We also observe that, as shown in Table \ref{tab:case1resultslr}, BiasCorr(probit, LR) and BiasCorr(probit, MLP) outperform Greene's method by 7.16\% and 22.79\%, respectively. As $\eta>1/(2-\Bar{s})$ for the Adult dataset, the result validates our theoretical comparison of BiasCorr and Greene's method. For the German and Drug datasets, we see that the value of $1/(2-\Bar{s})$ is not less than $\eta$. However, BiasCorr still outperforms Greene's method across all 4 combinations of settings for $g_s$ and $g_y$. This shows that our BiasCorr algorithm, which improves Greene's method by incorporating pseudolabel generation and a soft selection assignment on samples in $\mathcal{D}_u$, produces a more robust classifier against MNAR sample selection bias.

\begin{figure}[t]
    \centering
    \subfloat[Adult. $\Bar{s} =0.2957$ (probit) and $\Bar{s}=0.2571$ (logit).]{%
        \includegraphics[clip,width=0.9\columnwidth]{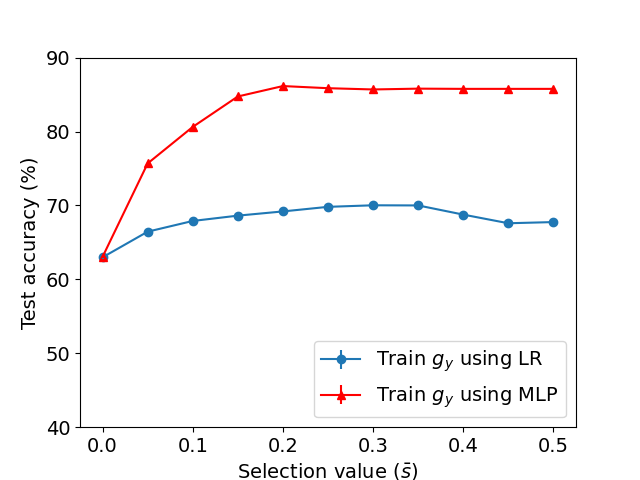}%
    }
   \vspace{-1.2em}
    \subfloat[German. $\Bar{s}=0.4240$ (probit) and $\Bar{s}=0.3957$ (logit).]{%
        \includegraphics[clip,width=0.9\columnwidth]{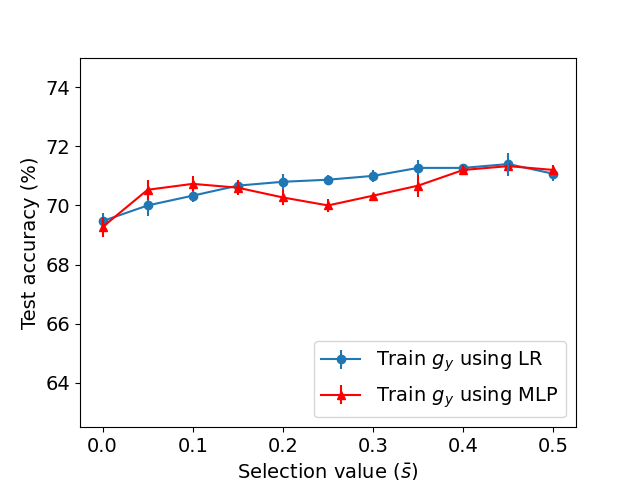}%
    }
    \caption{Evaluation of BiasCorr using different assignments of $\Bar{s}$ on samples in $\mathcal{D}_u$. Estimates of $\Bar{s}$ obtained after training $g_s$ are also given.}
    \label{fig:ablation1}
\end{figure}

\begin{table}[t]
    \small
    \centering
    \caption{Performance of baselines across different values of $\eta$ compared to BiasCorr using the Drug dataset.}
    \resizebox{0.47\textwidth}{!}{\begin{tabular}{|c|c||c|c|}
    \hline
       \textbf{Method} & $1/(2-\Bar{s})$ & Test Acc. (\%) & F1 Score (\%) \\
    \hline
    \hline
        \multicolumn{4}{|c|}{$\eta = 0.5$} \\
    \hline
    \hline
        SSBias & - & 65.72 $\pm$ 0.00 & 56.70 $\pm$ 0.00 \\
        Greene's method & - & 65.90 $\pm$ 0.27 & 55.67 $\pm$ 0.24 \\
        BiasCorr (probit, LR) & 0.6365 & \textbf{68.23 $\pm$ 0.38} & \textbf{62.77 $\pm$ 0.60} \\
        BiasCorr (LR, LR) & 0.6224 & 67.95 $\pm$ 0.52 & 61.97 $\pm$ 0.82 \\
    \hline
    \hline
        \multicolumn{4}{|c|}{$\eta = 0.6$} \\
    \hline
    \hline
        SSBias & - & 68.55 $\pm$ 0.00 & 62.61 $\pm$ 0.00 \\
        Greene's method & - & 67.63 $\pm$ 0.32 & 60.38 $\pm$ 0.51 \\
        BiasCorr (probit, LR) & 0.6279 & 69.40 $\pm$ 0.07 & 65.16 $\pm$ 0.06 \\
        BiasCorr (LR, LR) & 0.6167 & \textbf{69.43 $\pm$ 0.00} & \textbf{65.19 $\pm$ 0.00} \\
    \hline
    \hline
        \multicolumn{4}{|c|}{$\eta = 0.7$} \\
    \hline
    \hline
        SSBias & - & 68.37 $\pm$ 0.00 & 59.78 $\pm$ 0.00 \\
        Greene's method & - & 67.10 $\pm$ 0.44 & 56.92 $\pm$ 0.69 \\
        BiasCorr (probit, LR) & 0.6258 & 69.22 $\pm$ 0.13 & 64.92 $\pm$ 0.13 \\
        BiasCorr (LR, LR) & 0.6072 & \textbf{69.43 $\pm$ 0.16} & \textbf{65.05 $\pm$ 0.13} \\
    \hline
    \end{tabular}
    }
    \label{tab:missingnessratio}
\end{table}

\begin{table*}[h]
    \small
    \centering
    \caption{Performance of baselines compared to BiasCorr$^*$.}
    \begin{tabular}{|c|c|c|c|c|c|c|}
    \hline
        \multirow{2}{*}{\textbf{Methods}} & \multicolumn{2}{c|}{\textbf{Adult}} & \multicolumn{2}{c|}{\textbf{German}} & \multicolumn{2}{c|}{\textbf{Drug}} \\ \cline{2-7} 
        & Train Acc. (\%) & Test Acc. (\%) & Train Acc. (\%) & Test Acc. (\%) & Train Acc. (\%) & Test Acc. (\%) \\
    \hline
        \textit{RFLearn}$^1$ & 78.04 $\pm$ 0.00 & 69.68 $\pm$ 0.00 & 76.02 $\pm$ 0.00 & 69.67 $\pm$ 0.00 & 75.82 $\pm$ 0.00 & 65.02 $\pm$ 0.00 \\
        RBA & 77.69 $\pm$ 0.00 & 69.59 $\pm$ 0.00 & 75.84 $\pm$ 0.00 & 67.33 $\pm$ 0.00 & 75.82 $\pm$ 0.00 & 65.55 $\pm$ 0.00 \\
        BiasCorr$^*$ (probit, LR) & 87.10 $\pm$ 0.02 & 69.84 $\pm$ 0.07 & 80.57 $\pm$ 0.09 & 70.47 $\pm$ 0.34 & 87.70 $\pm$ 0.13 & \textbf{68.52 $\pm$ 0.14} \\
        BiasCorr$^*$ (LR, LR) & 87.37 $\pm$ 0.03 & 69.75 $\pm$ 0.02 & 80.57 $\pm$ 0.16 & \textbf{70.67 $\pm$ 0.30} & 87.98 $\pm$ 0.11 & 68.34 $\pm$ 0.21 \\
        BiasCorr$^*$ (probit, MLP) & 94.00 $\pm$ 0.35 & \textbf{85.75 $\pm$ 0.01} & 79.66 $\pm$ 0.21 & 70.07 $\pm$ 0.13 & 87.20 $\pm$ 0.10 & 68.23 $\pm$ 0.13 \\
        BiasCorr$^*$ (LR, MLP) & 93.78 $\pm$ 0.36 & 85.62 $\pm$ 0.02 & 79.40 $\pm$ 0.17 & 69.87 $\pm$ 0.16 & 87.40 $\pm$ 0.15 & 67.99 $\pm$ 0.26 \\
    \hline
    \end{tabular}
    \label{tab:case2results}
\end{table*}


\noindent \textbf{Execution Time.} We also report the execution times of training $h$ using Greene's method and BiasCorr in Table \ref{tab:execution}, where the experiments were conducted on the Dell XPS 8950 9020 with a Nvidia GeForce RTX 3080 Ti.  We see that BiasCorr trains slower than Greene's method for the Adult dataset while BiasCorr has a slightly faster execution time than Greene's method for the 
German and Drug datasets. For instance, BiasCorr(LR, LR) executes 6.09 seconds slower than Greene's method for the Adult dataset and 0.19 and 0.86 seconds faster than Greene's method for the German and Drug datasets, respectively. 


\noindent \textbf{Sensitivity Analysis.} We further conduct an empirical evaluation of the performance of BiasCorr when considering different assignments for the soft selection value $\Bar{s}$ on samples in $\mathcal{D}'_u$, up to $\Bar{s}=0.5$. In this particular study, we run Algorithm \ref{alg:cap} except we ignore the training of $g_s$ using the probit or logit model. Figure \ref{fig:ablation1} shows the results of this experiment over the Adult and German datasets. We see that when training $g_y$ under both logistic regression and an MLP, the performance of BiasCorr peaks within the range of the estimates we obtain by computing the average of predictions given by $g_s$ on samples in $\mathcal{D}_u$.

We also evaluate how modifying $\eta$ on the training set affects the performance of BiasCorr on the testing set. We look at the values of 0.5, 0.6, and 0.7 for $\eta$. We train our method using the Drug dataset for this experiment. Using testing accuracy and F1 score as evaluation metrics and logistic regression to train $g_y$, we report the results of this sensitivity analysis over the Drug dataset in Table \ref{tab:missingnessratio}. We first see that the $\eta > 1/(2-\Bar{s})$ condition is satisfied when $\eta = 0.7$. For $\eta=0.7$, BiasCorr(probit, LR) and BiasCorr(LR, LR) outperform SSBias and Greene's method based on test accuracy and F1 score, as indicated in Table \ref{tab:missingnessratio}. For the other two values of $\eta$, where the condition is not satisfied, BiasCorr(probit, LR) and BiasCorr(LR, LR) still outperform SSBias and Greene's method.

\subsection{Experiments on BiasCorr$^*$}


For the biased training set of labeled samples, we use the same set $\mathcal{D}_s$ that was used in the experiments on BiasCorr and leave the rest of the samples unlabeled as part of the set $\mathcal{D}_N$. We fix the number of samples $n$ drawn from $\mathcal{D}_N$ to be the number of samples that is obtained after splitting each dataset.

\noindent \textbf{Baselines.} We compare BiasCorr$^*$ to the following baselines that were proposed to learn classification under MAR sample selection bias where samples from the unbiased target distribution are unlabeled: (a) a robust non-fair version of \textit{RFLearn}$^1$ \cite{du2021robust}, which considers the empirical frequencies of each record in $\mathcal{D}_s$ and the unlabeled testing set to estimate the true probability of selection, and (b) the Robust Bias Aware (RBA) classifier \cite{liu2014robust}, which uses minimax estimation to learn against a worst-case conditional label distribution.

\noindent \textbf{Results.} As shown in Table \ref{tab:case2results}, BiasCorr$^*$, under all combinations of settings for $g_s$ and $g_y$, outperforms the baselines when trained on the Adult, German, and Drug datasets. For instance, BiasCorr$^*$(probit, MLP) outperforms RBA by 16.09\% for the Adult dataset in terms of testing accuracy. For the German dataset, BiasCorr$^*$(probit, MLP) has a testing accuracy that is 2.74\% higher than RBA. Moreover, the testing accuracy for BiasCorr$^*$ (probit, LR) is 3.14\% higher than RBA for the German dataset. These results suggest that BiasCorr$^*$ can outperform other classifiers trained under sample selection bias regardless of the type of model chosen for $g_s$ and $g_y$ or the proportion of unbiased, unlabeled samples in $\mathcal{D}_{aug}$.


\section{Conclusion}

In this paper, we have proposed a framework, BiasCorr, to learn a classifier that is robust against sample selection bias on the training dataset that causes a subset of training samples to have non-randomly missing labels. As a significant improvement to a formulation previously proposed to model MNAR sample selection bias, BiasCorr trains a robust classifier after learning separate classifiers to predict pseudolabels and estimate a soft selection value assignment for these samples. Theoretical analysis on the bias of BiasCorr provides a guarantee for this improvement based on the level of missingness in the training set. 
Experimental results on two real-world demonstrate not only the robustness of classifiers under this framework, but also their better performance than baselines. In the future, we plan to extend this framework to learn more complex non-linear regression models such as kernel ridge regression.




\bibliographystyle{ACM-Reference-Format}
\bibliography{arxiv}

\clearpage

\appendix

\section{Selection and Prediction Features} \label{sec:features}

\begin{table}[h]
    \small
    \centering
    \caption{Features used for selection/prediction. Prediction features are in italic font while selection features are in either italic or regular font.}
    \begin{tabular}{|l|l|}
    \hline
        \textbf{Dataset} & Features \\
    \hline
        Adult & \textit{Age}, \textit{Target}, \textit{Education-Num,} \\
            & \textit{Capital Gain, Capital Loss,} \textit{Hours per week}, \\
            & Country\_Canada, \textit{Occupation\_Adm-clerical},    \\
            & \textit{Occupation\_Armed-Forces}, \textit{Occupation\_Sales}, \\
            & \textit{Occupation\_Craft-repair}, \textit{Occupation\_Other-service},   \\
            & \textit{Occupation\_Prof-specialty}, \textit{Occupation\_Tech-support},  \\
            & \textit{Occupation\_Exec-managerial}, \\
            & \textit{Occupation\_Farming-fishing},  \\
            & \textit{Occupation\_Protective-serv}, \\
            & \textit{Occupation\_Machine-op-inspct}, \\
            & \textit{Occupation\_Priv-house-serv}, \\
            & \textit{Occupation\_Handlers-cleaners}, \\
            & Occupation\_Transport-moving, \textit{Marital Status}, \\
            & Relationship\_Other-relative, Relationship\_Husband,  \\
            & Relationship\_Wife, Relationship\_Not-in-family,  \\
            & Relationship\_Own-child, Relationship\_Unmarried \\
    \hline
        German & status checking, \textit{duration}, credit history, \\
            & credit amount, \textit{savings account}, \textit{telephone}, \\
            & \textit{last employment}, \textit{age}, \textit{personal status and sex}, \\
            & \textit{last residence}, property, \textit{existing credits}, \\
            & \textit{other plans}, \textit{liable}, \textit{foreign worker} \\
    \hline
        Drug & Age, \textit{Gender}, \textit{Education}, \textit{Country}, \textit{Ethnicity}, \\
            &  Nscore, \textit{Escore}, Oscore, Ascore, Cscore, Impulsive, SS \\
    \hline
    \end{tabular}
    \label{tab:features}
\end{table}

\section{Implementation Details} \label{sec: hyperparam}
We pre-process the Adult dataset by making the marital status attribute binary (married or not-married), adjusting the country attribute to where countries represented by at most 150 records are considered in the Other category, and deleting the final weight and race attributes.

Rather than keeping the number of training epochs fixed, BiasCorr, which is optimized using stochastic gradient descent with a learning rate of 0.01 and a weight decay of $1\times 10^{-4}$, keeps training until the percent change in loss is less than 0.025\% for the Adult dataset and 0.05\% for the German and Drug datasets. The prediction and selection coefficients $\bm{\beta}$ and $\bm{\gamma}$ are initialized to zero while $\sigma$ and $\rho$ are initialized to $0.01$. The number of random draws $R$ is set to 200. All baselines are implemented using Pytorch. \\

\section{Proof Details of Bias Analysis in Section \ref{sec: bias analysis}} \label{app: proof}
\subsection{Proof of Lemma \ref{lemma: bias_greene}}
\label{sec: bias_greene}
Following the definition, the bias of estimator $\hat{\mathcal{L}}$ from Greene's method is:
\[ \textit{Bias}(\hat{\mathcal{L}}) = |\mathcal{L}^* - \mathbb{E}_{\mathcal{D}_{tr}}[\hat{\mathcal{L}}] |\]
The loss function $\mathcal{L}^*$ on all training samples is defined as:
\begin{equation}
\mathcal{L}^* = -\frac{1}{|\mathcal{D}_{tr}|} \sum_{i \in \mathcal{D}_{tr}} \log P(y_i|\bm{x}_i )  =  -\frac{1}{n} \sum_{i=1}^n \log\left( f(y_i \vert \bm{x}_{i}^{(p)}) \right)     
\end{equation}
where $f(y_i \vert \bm{x}_{i}^{(p)})$ takes the form of logistic regression. By plugging in the expression of $\mathcal{L}^*$ and $\hat{\mathcal{L}}$ we have:

\begin{equation}
        \textit{Bias}(\hat{\mathcal{L}}) = \bigg |  -\frac{1}{n} \sum_{i=1}^n \log\left( f(y_i \vert \bm{x}_{i}^{(p)}) \right)   -  \mathbb{E}_{\mathcal{D}_{tr}}\bigg[ -\frac{1}{n}\sum_{i=1}^n \hat{l}_i \bigg] \bigg | 
\label{eq: bias}
\end{equation}

where 
\begin{equation}
\small
    \hat{l}_i = \log\bigg( \frac{1}{R}\sum_{r=1}^{R} [(1-s_i)+s_i f(y_i\vert \bm{x}_{i}^{(p)},\epsilon_{ir})] \\
    \cdot P(s_i\vert \bm{x}_{i}^{(s)}, \epsilon_{ir}) \bigg)
\label{eq: l_estimate}
\end{equation}

Denoting the expectation of selection model $P(s_i\vert \bm{x}_{i}^{(s)}, \epsilon_{ir})$ and prediction model $f(y_i\vert \bm{x}_{i}^{(p)},\epsilon_{ir})$ over $R$ random draws on the error terms as $\hat{p}(s_i),  \hat{f}(y_i\vert \bm{x}_{i}^{(p)})$ respectively, the log-likelihood function for each tuple $i$ in Eq. (\ref{eq: l_estimate}) can be further rewritten as: 

\[\hat{l}_i = \log \bigg( (1-s_i) \cdot \hat{p}(s_i) + s_i \cdot  \hat{f}(y_i\vert \bm{x}_{i}^{(p)}) \cdot  \hat{p}(s_i)   \bigg) \]

After taking the expectation over all the training samples, the latter term in Eq. (\ref{eq: bias}) could be expressed as:

\begin{equation}
\begin{split}
&\mathbb{E}_{\mathcal{D}_{tr}}\bigg[ -\frac{1}{n}\sum_{i=1}^n \hat{l}_i \bigg] \\  
=& -\frac{1}{n}\sum_{i=1}^n \mathbb{E}_{\mathcal{D}_{tr}} \bigg[\log \bigg( (1-s_i) \cdot \hat{p}(s_i) + s_i \cdot  \hat{f}(y_i\vert \bm{x}_{i}^{(p)}) \cdot  \hat{p}(s_i)   \bigg)    \bigg]\\
=&-\frac{1}{n}\sum_{i=1}^n  \log \bigg(   (1- p(s_i)) \cdot \hat{p}(s_i) + p(s_i) \cdot  \hat{f}(y_i\vert \bm{x}_{i}^{(p)}) \cdot \hat{p}(s_i) \bigg)     \\
=& -\frac{1}{n}\sum_{i=1}^n  \log \bigg( \hat{p}(s_i) + p(s_i)\hat{p}(s_i)(  \hat{f}(y_i\vert \bm{x}_{i}^{(p)}) -1 )   \bigg)  
\end{split}
\end{equation}

Finally, plugging the results above into Eq. (\ref{eq: bias}) leads to the bias of  $\hat{\mathcal{L}}$:
\begin{equation}
\textit{Bias}(\hat{\mathcal{L}}) = \bigg |  \frac{1}{n}\sum_{i=1}^n  \log \bigg( \frac{ f(y_i\vert \bm{x}_{i}^{(p)})}{  \hat{p}(s_i) + p(s_i)\hat{p}(s_i)(  \hat{f}(y_i\vert \bm{x}_{i}^{(p)}) -1 ) }  \bigg)  \bigg|
\end{equation}

\subsection{Proof of Lemma \ref{lemma: bias_biascorr}}

Similar to the proof in Section \ref{sec: bias_greene} we derive the bias of the loss function for BiasCorr Algorithm. We plug in the expression of $\mathcal{L}^*$ and $\hat{\mathcal{L}}'$ and obtain the following equation:

\begin{equation}
        \textit{Bias}(\hat{\mathcal{L}}') = \bigg |  -\frac{1}{n} \sum_{i=1}^n \log\left( f(y_i \vert \bm{x}_{i}^{(p)}) \right)   -  \mathbb{E}_{\mathcal{D}_{tr}}\bigg[ -\frac{1}{n}\sum_{i=1}^n \hat{l}'_i \bigg] \bigg | 
\label{eq: bias_corr}
\end{equation}
where
\begin{equation}
\hat{l}'_i  =   \log\bigg( \frac{1}{R}\sum_{r=1}^{R} [(1-s'_i)+s'_if(y'_i\vert \bm{x}_{i}^{(p)},\epsilon_{ir})] \\
    \cdot P(s'_i\vert \bm{x}_{i}^{(s)}, \epsilon_{ir}) \bigg)
\end{equation}
 
Using the similar derivation procedure in Section \ref{sec: bias_greene}, the latter term in Eq. (\ref{eq: bias_corr}) could be further calculated as :
\begin{equation}
\begin{split}
&\mathbb{E}_{\mathcal{D}_{tr}}\bigg[ -\frac{1}{n}\sum_{i=1}^n \hat{l}'_i \bigg] \\
=&  -\frac{1}{n}\sum_{i=1}^n  \mathbb{E}_{\mathcal{D}_{tr}}\bigg[ \log \bigg( \frac{1}{R}\sum_{r=1}^{R} [(1-s'_i)+s'_i \hat{f}(y'_i\vert \bm{x}_{i}^{(p)})] \cdot \hat{p}(s'_i) \bigg)\bigg] \\
=&-\frac{1}{n}\sum_{i=1}^n  \log \bigg(   (1- p(s_i) - \Bar{s}\eta) \cdot \hat{p}(s'_i) + \\
&\;\;\;\;\;\;\;\;\;\;\;\;\;\;\;\;\;\;\;\;  (p(s_i)+\Bar{s}\eta) \cdot \hat{f}(y'_i\vert \bm{x}_{i}^{(p)}) \cdot \hat{p}(s'_i) \bigg)     \\
=& -\frac{1}{n}\sum_{i=1}^n  \log \bigg( \hat{p}(s'_i) + (p(s_i)+ \Bar{s}\eta )\hat{p}(s'_i)( \hat{f}(y'_i\vert \bm{x}_{i}^{(p)}) -1 )   \bigg)
\end{split}
\end{equation}
Finally, the bias regarding to the loss function of BiasCorr algorithm is listed as follows:
\begin{equation}
      \textit{Bias}(\hat{\mathcal{L}}') =  \bigg |  \frac{1}{n}\sum_{i=1}^n  \log \bigg( \frac{ f(y_i\vert \bm{x}_{i}^{(p)})}{  \hat{p}(s'_i) + (p(s_i)+ \Bar{s}\eta)\hat{p}(s'_i)( \hat{f}(y'_i\vert \bm{x}_{i}^{(p)}) -1 ) }  \bigg)  \bigg|
\end{equation}

From the result above we can see even the estimated selection variable model and outcome model are accurate, that is, $\hat{p}(s_i) = p(s_i)$ and $  \hat{f}(y_i\vert \bm{x}_{i}^{(p)}) = f(y_i\vert \bm{x}_{i}^{(p)})$, the biases of both loss function estimators $\hat{\mathcal{L}}$ and $\hat{\mathcal{L}'}$ are still non-zero.

\subsection{Proof of Theorem \ref{thm: bias_diff}}
To fairly compare the biases from two loss function estimators, we assume that $\hat{p}(\cdot)$ accurately estimate the true selection model $p(s_i)$, that is, $\hat{p}(s_i) = p(s_i)$ for both Greene's method and BiasCorr algorithm. We also assume that $\hat{f}(y_i\vert \bm{x}_{i}^{(p)})=\hat{f}(y'_i\vert \bm{x}_{i}^{(p)})$, i.e. the outcome models for Greene's method and BiasCorr are exactly the same.
The two biases thus reduce to the following form:

\begin{equation}
\textit{Bias}(\hat{\mathcal{L}}) = \bigg |  \frac{1}{n}\sum_{i=1}^n  \log \bigg( \frac{ f(y_i\vert \bm{x}_{i}^{(p)})}{  p(s_i) + p(s_i)^2( \hat{f}(y_i\vert \bm{x}_{i}^{(p)}) -1 ) }  \bigg)  \bigg|
\end{equation}

\begin{equation}
\textit{Bias}(\hat{\mathcal{L}'}) = \bigg |  \frac{1}{n}\sum_{i=1}^n  \log \bigg( \frac{ f(y_i\vert \bm{x}_{i}^{(p)})}{  p(s_i) + \Bar{s}\eta + (p(s_i)+\Bar{s}\eta)^2( \hat{f}(y_i\vert \bm{x}_{i}^{(p)}) -1 ) }  \bigg)  \bigg|
\end{equation}

We first define the difference between the denominators inside the log-likelihood function for each tuple $i$ as follows:

\begin{equation}
\begin{split}
\textit{DIFF(i)} = &\bigg[ p(s_i) + \Bar{s}\eta + (p(s_i)+\Bar{s}\eta)^2( \hat{f}(y_i\vert \bm{x}_{i}^{(p)}) -1 ) \bigg] - \\
&\bigg[  p(s_i) + p(s_i)^2( \hat{f}(y_i\vert \bm{x}_{i}^{(p)}) -1 )  \bigg]   \\
=& \Bar{s}\eta + (\hat{f}(y_i\vert \bm{x}_{i}^{(p)}) -1)[(p(s_i)+\Bar{s}\eta)^2 - p(s_i)^2]\\
=& \Bar{s}\eta+ (\hat{f}(y_i\vert \bm{x}_{i}^{(p)}) -1)(2p(s_i) + \Bar{s}\eta) \cdot \Bar{s}\eta\\
=& \Bar{s}\eta \cdot [1 + (\hat{f}(y_i\vert \bm{x}_{i}^{(p)}) -1)(2p(s_i) + \Bar{s}\eta)]
\end{split}
\label{eq: diff}
\end{equation}






Since $\mathcal{L}^*$ denotes the minimized loss function with the maximized likelihood function, we have:

\begin{equation}
\begin{split}
 &\textit{Bias}(\hat{\mathcal{L}})- \textit{Bias}(\hat{\mathcal{L}'}) = \bigg( \mathbb{E}_{\mathcal{D}_{tr}}[\hat{\mathcal{L}}] - \mathcal{L}^*  \bigg) - \bigg( \mathbb{E}_{\mathcal{D}_{tr}}[\hat{\mathcal{L}'}] - \mathcal{L}^* \bigg)\\
=&\bigg( \mathbb{E}_{\mathcal{D}_{tr}}\bigg[ -\frac{1}{n}\sum_{i=1}^n \hat{l}_i \bigg] + \frac{1}{n} \sum_{i=1}^n \log\left( f(y_i \vert \bm{x}_{i}^{(p)}) \right)  \bigg) -  \\
&\bigg( \mathbb{E}_{\mathcal{D}_{tr}}\bigg[ -\frac{1}{n}\sum_{i=1}^n \hat{l}'_i \bigg] +\frac{1}{n} \sum_{i=1}^n \log\left( f(y_i \vert \bm{x}_{i}^{(p)}) \right) \bigg) \\
=& \frac{1}{n}\sum_{i=1}^n  \log \bigg(  p(s_i) + \Bar{s}\eta + (p(s_i)+\Bar{s}\eta)^2( \hat{f}(y_i\vert \bm{x}_{i}^{(p)}) -1 )  \bigg)\\
&- \frac{1}{n} \sum_{i = 1}^n \log\bigg(  p(s_i) + p(s_i)^2( \hat{f}(y_i\vert \bm{x}_{i}^{(p)}) -1 )\bigg)  \geq \frac{1}{n}\sum_{i=1}^n \textit{DIFF(i)} \\
\end{split}
\end{equation}

As defined in Eq. (\ref{eq: diff}), we further have:

\begin{equation}
\begin{split}
&  \frac{1}{n}\sum_{i=1}^n \textit{DIFF(i)}\\
=&  \Bar{s}\eta \cdot \frac{1}{n} \sum_{i=1}^n \big[1 + (\hat{f}(y_i\vert \bm{x}_{i}^{(p)}) -1)(2p(s_i) + \Bar{s}\eta) \big] \\ 
=& \Bar{s}\eta \cdot \bigg[ 1 + \frac{1}{n} \sum_{i=1}^n \big[ (\hat{f}(y_i\vert \bm{x}_{i}^{(p)}) -1)(2p(s_i) + \Bar{s}\eta) \big] \bigg]\\
=& \Bar{s}\eta \cdot \bigg[ 1 -  \Bar{s}\eta +  \frac{1}{n} \sum_{i=1}^n \big[ 2\hat{f}(y_i\vert \bm{x}_{i}^{(p)}) p(s_i) - 2p(s_i) + \hat{f}(y_i\vert \bm{x}_{i}^{(p)}) \cdot \Bar{s}\eta\big] \bigg] \\ 
=& \Bar{s}\eta \cdot \bigg[  \underbrace{1 - \Bar{s}\eta - \frac{1}{n} \sum_{i=1}^n 2p(s_i)}_\text{term 1} +  \underbrace{\big[ \frac{1}{n} \sum_{i=1}^n \hat{f}(y_i\vert \bm{x}_{i}^{(p)}) (2p(s_i) + \Bar{s}\eta)}_\text{term 2} \big] \bigg]
\end{split}
\end{equation}
Since both $\hat{f}(y_i\vert \bm{x}_{i}^{(p)})$ and $p(s_i)$ lie in $(0,1)$ for each tuple $i$, term 2 is still a positive value after summation and averaging over all the $n$  training tuples. For term 1,  notice that  $\frac{1}{n} \sum_{i=1}^n p(s_i) = \frac{m}{n}$ since $p(\cdot)$ represents the ground truth selection model. In order to ensure term 1 is larger than 0 we need to satisfy:

\begin{equation}
 1 - \Bar{s}\eta - 2(1-\eta) > 0   
\end{equation}
Solving the Equation above leads to:
\begin{equation}
\eta > \frac{1}{2 - \bar{s}}
\end{equation}

Thus, if the ratio $\eta = |\mathcal{D}_u| / |\mathcal{D}_{tr}| =  1 - m/n$ is larger than $1/(2 - \bar{s})$, we obtain the result $\textit{Bias}(\hat{\mathcal{L}})- \textit{Bias}(\hat{\mathcal{L}'}) > 0$, and our BiasCorr method achieves lower bias compared to Greene's method.

\end{document}